\documentclass[10pt,twocolumn,letterpaper]{article}

\usepackage{iccv}
\usepackage{times}
\usepackage{epsfig}
\usepackage{graphicx}
\usepackage{amsmath}
\usepackage{amssymb}
\usepackage{authblk}
\usepackage{amsthm}
\usepackage{appendix}
\usepackage{xcolor, colortbl}
\definecolor{Gray}{gray}{0.7}
\definecolor{Gray1}{gray}{0.9}
\usepackage{pifont}
\usepackage{subfig}
\usepackage{multirow}
\usepackage[ruled,vlined,linesnumbered]{algorithm2e}
\newtheorem{asu}{Observation}
\usepackage[font=small]{caption}

\usepackage{subfig}

\usepackage[pagebackref=true,breaklinks=true,letterpaper=true,colorlinks,bookmarks=false]{hyperref}

\iccvfinalcopy 

\def\iccvPaperID{2934} 

\ificcvfinal\fi

\begin{document}

\title{PoseFace: Pose-Invariant Features and Pose-Adaptive Loss\\ for Face Recognition}

\author{
  \text{\qquad \qquad}
  Qiang Meng\thanks{Algorithm Research, AiBee Inc.}\quad
  Xiaqing Xu$^*$\quad
  Xiaobo Wang\thanks{Institute of Automation, Chinese Academy of Sciences}\quad
  Yang Qian \thanks{The University of Sydney}
  \quad
  Yunxiao Qin \thanks{Northwestern Polytechnical University}\quad
  \text{\qquad\qquad }  
  Zezheng Wang$^*$\quad
  Chenxu Zhao\thanks{Corresponding author} \thanks{Academy of Sciences, Mininglamp Technology}\quad
  Feng Zhou$^*$\quad
  Zhen Lei \thanks{CBSR \& NLPR, Institute of Automation, Chinese Academy of Sciences, Beijing, China}\ \ \ \thanks{School of Artificial Intelligence, University of Chinese Academy of Sciences}
  }


\maketitle

\begin{abstract}
 Despite the great success achieved by deep learning methods in face recognition, severe performance drops are observed for large pose variations in unconstrained environments (\textit{e.g.}, in cases of surveillance and photo-tagging).
To address it, current methods either deploy pose-specific models or frontalize faces by additional modules. Still, they ignore the fact that identity information should be consistent across poses and are not realizing the data imbalance between frontal and profile face images during training. In this paper, we propose an efficient PoseFace framework which utilizes the facial landmarks to disentangle the pose-invariant features and exploits a pose-adaptive loss to handle the imbalance issue adaptively. Extensive experimental results on the benchmarks of Multi-PIE, CFP, CPLFW and IJB have demonstrated the superiority of our method over the state-of-the-arts.
\end{abstract}

\section{Introduction}

\begin{figure}[t]
    \centering
    \includegraphics[width=0.48\textwidth]{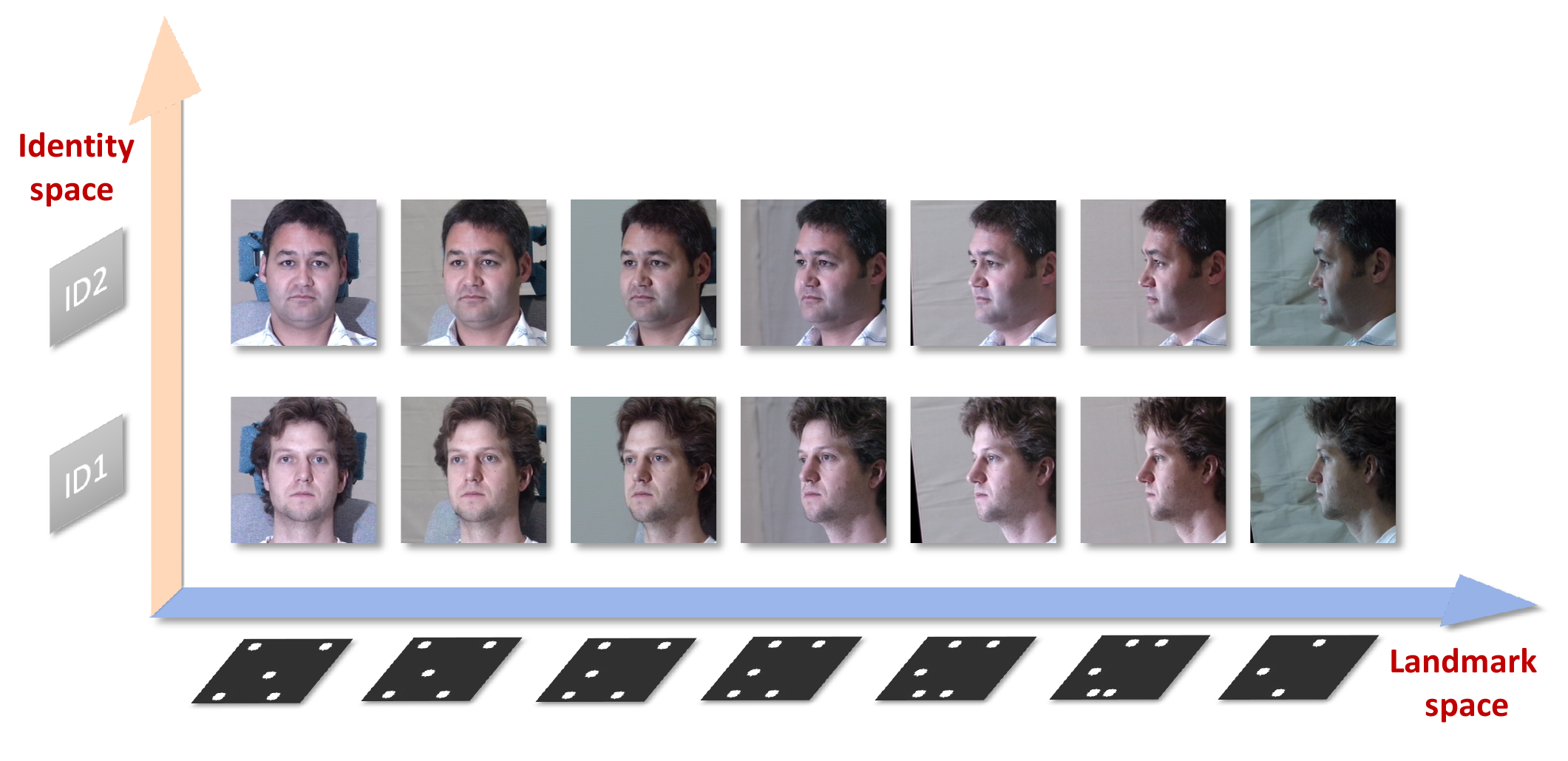}
    \caption{Face images are mainly composed of identity information and pose information.
      As pose information is encoded in the facial landmarks, we decompose the faces into identities and landmarks, which distribute in two orthogonal subspaces separately.}
    \label{fig:intro}
  \end{figure}{}
  
Recent years have witnessed the breakthrough of deep convolutional face recognition techniques~\cite{du2020elements, schroff2015facenet,sohn2016improved,deng2019arcface,wang2018support,xu2020searching, meng2021magface}.
However, face recognition with large pose variations is still a great challenge in computer vision.
The performance of most face recognition models degrades by over 10\% from frontal-frontal to frontal-profile recognition~\cite{sengupta2016frontal}.
That is mainly because: 
(a) proportions of frontal and profile faces in most modern datasets are highly imbalanced~\cite{cao2018pose} and
that prevents data-driven deep models to capture robust identity information;
(b) profile faces are intrinsically hard for recognition.
A face image mainly consists of identity information and pose information.
How to efficiently disentangle them is still an open issue.

To recognize faces in large poses, existing methods can be typically categorized into two classes.
The first one is to normalize profile faces into a frontal view~\cite{qian2019unsupervised,zhao2018towards,tran2017disentangled,hu2018pose,huang2017beyond,yin2017towards}.
Besides the artifacts, synthesizing faces also suffers from extra computations as well as loss of identity information, limiting their usage for recognition.
The second category aims at learning pose-invariant facial features.
Early works~\cite{masi2016pose,yin2017multi,xiong2015conditional} train multiple pose-specific models for faces under various poses.
To simplify the process, DFN~\cite{he2019deformable} and DREAM~\cite{cao2018pose} unify multiple paths into a pose-aware transformation over features.
However, their disentanglements are incomplete and not straightforward, as well as cannot be applied to in-the-wild recognition datasets where simultaneous existences of frontal and profile faces for each class cannot be guaranteed.


To address the aforementioned issues, we propose a novel and efficient framework called PoseFace which contributes to large-pose face recognition in two aspects.
The first one is to achieve pose-invariant embeddings in an end-to-end and self-disentanglement manner, instead of naively comparing features from frontal and profile faces.
Specifically,  we map each input face into pose and identity features simultaneously, and enforce orthogonality between subspaces of two features as shown in Figure~\ref{fig:intro}.
To achieve complete disentanglement, we
(a) constrain mappings to two subspaces to be linear and orthogonal;
(b) ensure the purity of pose features (\ie, only relates to pose) and therefore avoid identity information to be disentangled into pose features.
To this end, we design an AutoEncoder which is pre-trained and builds one-to-one mappings between facial landmarks and feature vectors.
The feature vectors are thus only related to landmarks and treated as pseudo labels for pose features.
With the supervision of such labels, identity information are kept for recognition to the most extent and that prevents disentanglement from decreasing recognition performances (indicated by our experiments in table~\ref{table:abl_ijb}).
In a word, the proposed self-disentanglement method is straightforward, and filters pose information efficiently while avoids wastage of identity information.

The second part of PoseFace is the pose-adaptive loss which handles data imbalances between profile and frontal faces.
Compared to frontal faces, profile faces are harder to be captured and processed (\eg, face detection and face alignment).
Therefore, profile faces (pitch/yaw/roll angles $>60^\circ$) are of small proportions in large and in-the-wild datasets (\eg, 0.19\% for MS1Mv2~\cite{deng2019arcface} according to our analysis).
Inspired by the weight-adaptive methods such as focal loss~\cite{lin2017focal}, Adacos~\cite{zhang2019adacos} and Adaptiveface~\cite{liu2019adaptiveface}, we down-weight losses for frontal faces by reformulating the prevailing ArcFace loss~\cite{deng2019arcface}.
In that way, we focus training on hard examples (\ie, large pose faces) and prevent the vast number of easy samples (\ie, near-frontal faces) from overwhelming the network.
To sum up, the main contributions of this paper are as follows:
\begin{enumerate}
  \itemsep0em
  \item We propose a self-disentanglement method which distributes identity and pose features into two orthogonal subspaces, and therefore obtain pose-invariant features.
    Our self-disentanglement doesn't require pairwise inputs and works in an end-to-end manner.
    Moreover, disentangled pose features are irrelevant to identities that avoids affecting recognitions.
  \item A pose-adaptive loss is designed to deal with the data imbalance problem.
    The loss increases penalties assigned to profile faces, and therefore prevents frontal faces from overwhelming the network.
  \item Extensive experiments are conducted on benchmarks for face recognition under large pose variations.
    The results have verified the superiority of our approach over current state-of-the-art methods.
    For example, PoseFace achieves human-level performances when recognizing the profile faces on the CFP benchmark. On the MultiPIE benchmark, our PoseFace achieves a 0.85\%,  4.87\%, 6.51\% boosts on the rank-1 recognition rates when identifying faces with yaw angle $60^\circ, 75^\circ, 90^\circ$ respectively.
  \end{enumerate}

\section{Related Works}

In this section, we first briefly review face recognition methods under large pose variations, which are mainly categorized into pose-invariant feature representation (Sec.~\ref{sec:relatedworks_1}) and face frontalization (Sec.~\ref{sec:relatedworks_3}).
Then, we list disentanglement methods used in face-related literatures and analyze differences between them and our self-disentanglement method in Sec.~\ref{sec:relatedworks_3}.

\subsection{Pose-Invariant Feature Representation}\label{sec:relatedworks_1}
Pose-invariant feature representation methods aim at exploring features that are robust to changes of poses.
One possible solution is the divide and conquer, \ie, using multiple pose-specific models to handle different-pose faces.
For example, c-CNN~\cite{xiong2015conditional} introduces dynamically activated kernels to form various network structures for different inputs.
Kernels in each layer are sparsely activated conditioned on present feature representations.
PAMs~\cite{masi2016pose} fuses scores from multiple pose-specific models to tackle pose variation.
p-CNN~\cite{yin2017multi} proposes a multi-task convolutional neural network where identity classification is the main task and pose, illumination, and expression estimations are the side tasks.
The efficiency issue of such multi-model frameworks limits their usage in many real-world applications.

Besides the divide and conquer, some works present a single model to learn pose-invariant features.
For example, DFN~\cite{he2019deformable} aligns features across different poses by exploiting deformable convolutional modules.
Their method takes paired samples from the same identity and penalizes the differences to reduce the intra-class feature variation.
DREAM~\cite{cao2018pose} adaptively adds residuals to the input deep representation to transform a profile face representation to a canonical frontal pose.
Conventional multi-view subspace approaches~\cite{sharma2012robust, sharma2012generalized,zhu2013deep} learn complex nonlinear transformations that respectively project images captured under different poses to the common space, where the intra-class variation is minimized.
These methods either cannot achieve complete disentanglement or fail to handle data imbalance issue.
In contrast, our proposed self-disentanglement method can be directly applied to datasets even with small proportions of profile faces.
The profile faces are further increase-weighted by a pose-adaptive loss.

\subsection{Face Frontalization}\label{sec:relatedworks_2}
Face frontalization methods normalize faces to a canonical frontal view and then use synthesized faces for recognition.
Recently, generative adversarial networks (GAN) has shown its great potential in synthesizing faces~\cite{qian2019unsupervised,zhao2018towards,tran2017disentangled,hu2018pose,huang2017beyond,yin2017towards,zhou2020rotate,cao2018learning}.
TP-GAN~\cite{huang2017beyond} and PIM~\cite{zhao2018towards} normalize faces by simultaneously perceiving global structures and local details.
FNM~\cite{qian2019unsupervised} is an unsupervised face normalization method which includes a face expert network and face attention discriminators.
In CAPG-GAN~\cite{hu2018pose}, head pose information not only guides the generator in the process, but also is used as a controllable condition during inference.
Despite the improvements achieved by GAN, it suffers from unreal textures and loss of identity information in synthesized faces, as well as its high computational cost.

\begin{figure*}[htb!]
    \centering
    \includegraphics[width=0.8\textwidth]{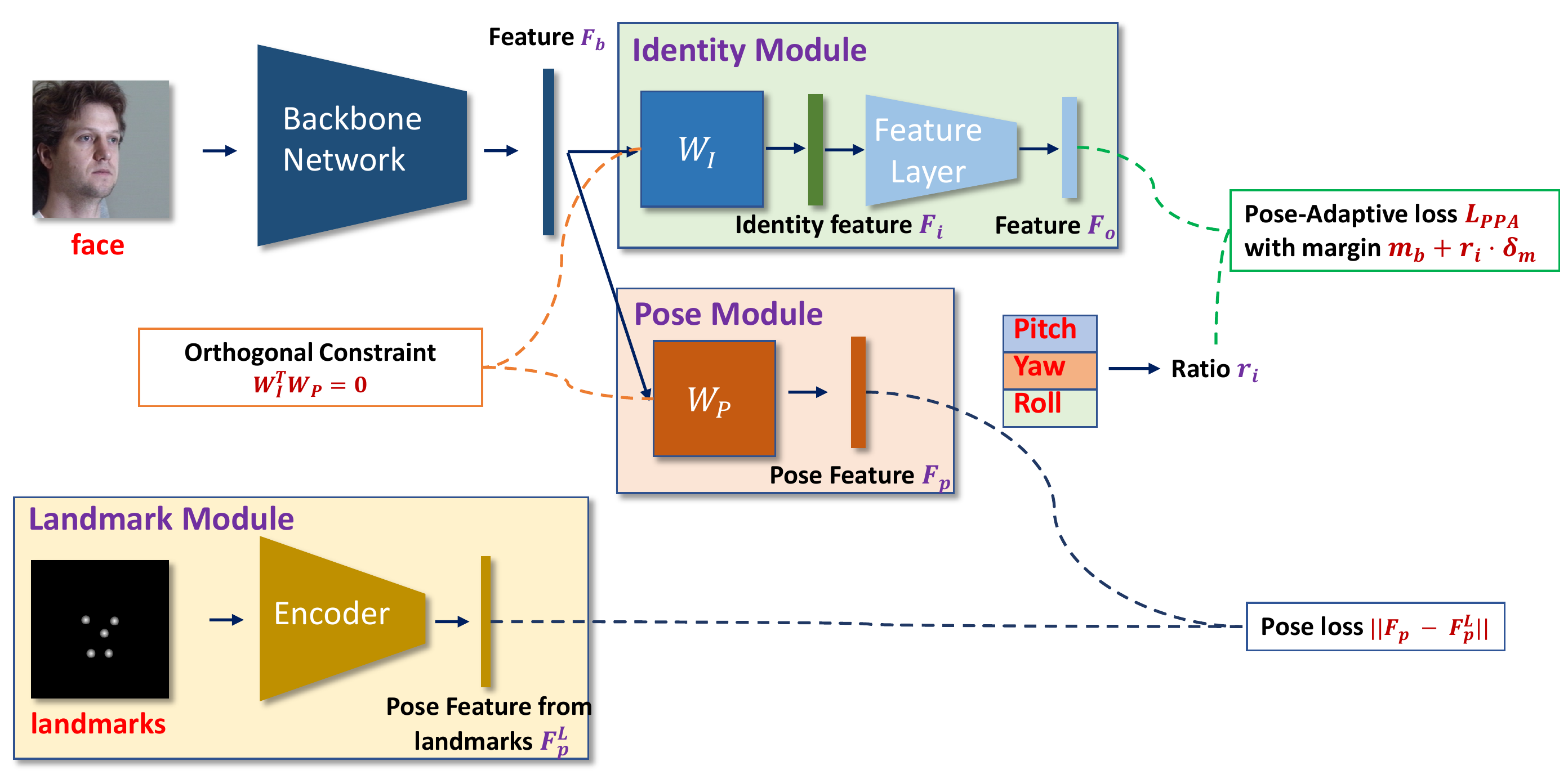}
    \caption{An overview of our proposed PoseFace framework. The pseudo label of pose feature $F_p^{L}$ is generated by a pre-trained Encoder network from facial landmarks.  The identity feature $F_i$ and pose feature $F_p$ are estimated by two fully connected layers with orthogonal weights $\mathbf W_I$ and $\mathbf  W_P$ after the backbone network. The network is trained by  (1) the pose-adaptive ArcFace loss with variable margins controlled by ratio $r_i$, which is estimated from facial angles, (2) the pose loss $\|F_p - F_p^{L}\|$ and (3) the orthogonal constraint $\mathbf W^{T}_I\mathbf W_P=0$.    }
    \label{fig:poseface}
\end{figure*}{}

\subsection{Disentanglement in Face Recognition}\label{sec:relatedworks_3}
A great amount of works for robust face recognition share the same high-level idea of disentanglement. 
By disentangling factors such as pose, age and expressions, they finally achieve face representations which are robust to factor changes.
Most works adopt generative adversarial networks or AutoEncoders, to either frontalize faces~\cite{huang2017beyond,cao2019towards,liu2018disentangling} or re-construct faces under different factors~\cite{cao20193d,tran2017disentangled, tran2019disentangling, wang2019adversarial}.
Besides non-trivial training process and extra computations/modules, these methods usually require extra information such as facial texture maps~\cite{cao2019towards}, 3D Morphable Model~\cite{cao20193d,liu2018disentangling,wang2019adversarial} and faces from multiviews~\cite{tran2017disentangled, tran2017disentangled}.
Most of them are based on theory and rarely applied to actual scenarios, according to the recent survey~\cite{ning2020multi}.

Wang \etal~\cite{wang2018orthogonal} presents another way of disentanglement by regressing ages from feature magnitudes. 
Even though naturally orthogonal to vector direction,  magnitude is a scalar and therefore unable to describe complex factors.
Other disentanglement methods penalize differences between faces from same identity but with different factors~\cite{he2019deformable,peng2017reconstruction} or enforce consistent distributions across modalities~\cite{yi2015shared, he2018wasserstein}.
These methods require pair-wise inputs from same identities and that restrict their usage in many scenarios, as profile and frontal faces are highly imbalance in modern recognition datasets.

Compared to previous works, our PoseFace is an efficient and straightforward framework.
The proposed self-disentanglement is achieved by mapping faces into identity and landmark spaces, which are irrelevant to each other by orthogonal constraints.
The whole process requires no pair-wise inputs or extra labels, and therefore can be trained in common datasets. 
In addition, the training speed and GPU memory consumptions are all close to the state-of-the-art ArcFace~\cite{deng2019arcface} as indicated in table~\ref{table:speed}.
During inference, PoseFace has no difference with ArcFace as only the identity module (in figure ~\ref{fig:poseface}) is needed.

\section{Methodology}

Our work is mainly based on the following two observations:
\begin{asu}
 The appearance of a face is composed of identity information and variation information (e.g., lighting, poses, and expressions)~\cite{wu2019disentangled,song2019occlusion}. Identity information and variation information should be irrelevant.  \label{assp:1}
\end{asu}
\begin{asu}
 Head pose information can be encoded by facial landmarks~\cite{hu2018pose}. Other information such as lighting and textures from the original face images are not that necessary for pose estimation. \label{assp:2}
\end{asu}

Note that in observation 2, 3D facial landmarks are normally needed to fully express out-of-plane 3D head poses.
2D facial landmarks are mostly distributed in a same plane and therefore insufficient to represent 3D head poses.
However, the key idea of PoseFace is to remove pose-related information out of 2D images by the self-disentanglement.
Therefore, we can employ 2D facial landmarks to help filter pose information in images.

The overall pipeline of the proposed method is shown in figure~\ref{fig:poseface}.
Our proposed PoseFace mainly consists of a landmark module, an identity module and a pose module.
The landmark module is pre-trained and served as the pose feature extractor.
During the training phase, the orthogonal constraint and pose loss are introduced to train the identity module and the pose module and achieve self-disentanglement.
The adaptive recognition loss is adjusted from the prevailing ArcFace~\cite{deng2019arcface} to increase-weight profile faces.
During testing, only the identity module is used for evaluation, same as ArcFace.
Below we will introduce our PoseFace from the perspectives of network details, pose-adaptive ArcFace (PAA) loss and pose-invariant features.

\subsection{Network Details}

As shown in figure~\ref{fig:poseface}, 
faces are first fed into the backbone network and then the identity/pose module disentangles backbone features to identity/landmark subspace.
The disentanglement is achieved by an orthogonal constraint over the linear layer in each module.
The landmark module takes landmarks as inputs and generates pseudo labels of pose features to supervise the learning of pose module.

\noindent\textbf{Pose Module.} The pose module is a linear layer with weight $\mathbf{W}_P$  which projects the backbone feature $F_b$ to the pose feature $F_p$.

\noindent\textbf{Identity Module.}
In the identity module, $F_b$ is first projected to identity feature $F_i$ by $W_I$ and then transferred to $F_o$ for recognition via an extra feature layer.
The feature layer, which takes the form of a fully connected layer in our implementation, is designed for the following two reasons:
(1) The projected feature $F_i$ may not be suitable for recognition and the extra layer can refine it for the task.
(2) The projection is defined on Euclidean space where the product of two vectors equals 0 (\ie, $v_1^Tv_2=0$) indicates orthogonal property in the space. In contrast, features for recognition are measured by cosine similarity. Therefore, $v_1^Tv_2=0$ means the two vectors are not similar. The domain difference is relieved 
by the introduced feature layer.

\noindent\textbf{Landmark Module.}
In PoseFace,  pose feature $F_p^L$ is directly estimated from facial landmarks and therefore is pose-related only.
With the supervision of such $F_P^L$ and an orthogonal constraint, pose information flows into the pose module while other information is well-kept in the identity module.
That accomplishes the disentanglement without hurting recognition performance (ablation study in section~\ref{sec:ablation} reveals the efficacy of such design). 

\begin{figure}[htb!]
    \centering
    \includegraphics[width=0.45\textwidth]{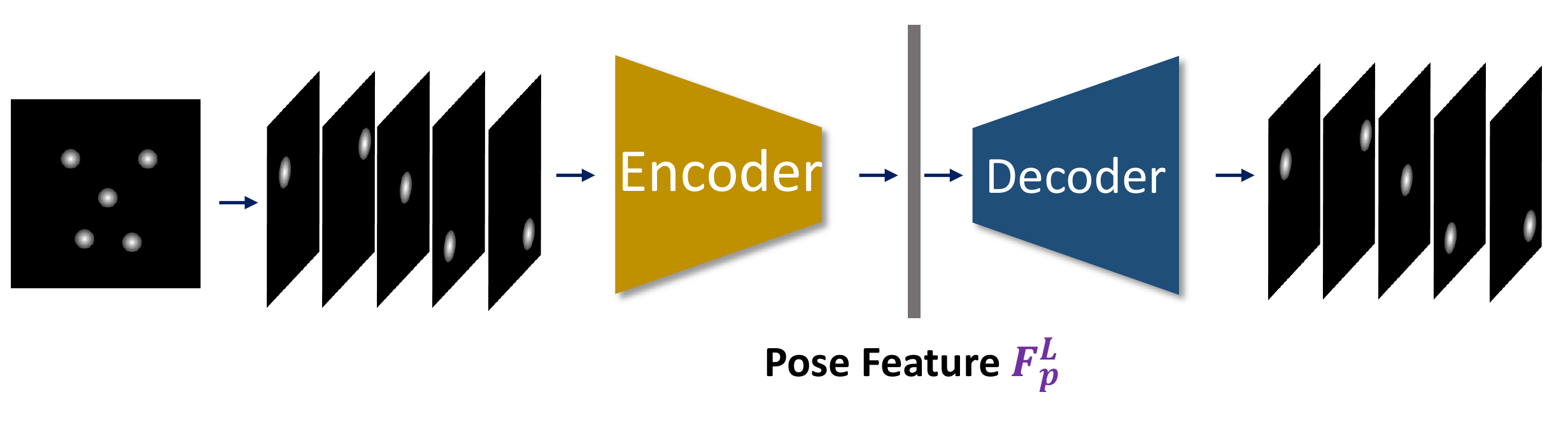}
    \caption{The AutoEncoder for extracting pose features. Heatmaps are synthesized from landmarks and we stack all heatmaps as inputs and outputs of the Auto-Encoder. }
    \label{fig:AEnet}
\end{figure}

To find the one-to-one mapping between pose feature and facial landmarks, we implement an Auto-Encoder network as shown in figure~\ref{fig:AEnet}.  The pre-trained Encoder is exploited to extract features in the landmark module. According to observation~\ref{assp:2}, we treat extracted features as pose features and use them as pseudo labels to train the pose module.  Considering the sparsity of heatmaps, we adopt a weighted $l_2$ loss to train the Auto-Encoder as shown in Eq.~\eqref{eq:ae}.
 \begin{equation}
   \small
  L_{ae}= \lambda_h\cdot \|\mathbf{H}_i\circ (\mathbf{H}_i-\mathbf{H}_o) \|_F  +  \|(\mathbf{1}-\mathbf{H}_i) \circ (\mathbf{H}_i-\mathbf{H}_o) \|_F
    \label{eq:ae}
\end{equation}
where $\|\cdot\|_F$ means Frobenius norm and $\circ$ means Hadamard product. $\mathbf{H}_i,\mathbf{H}_o $ are input and output heatmaps, respectively. $ \lambda_h$ is the weight of the loss assigned to the area where $\mathbf{H}_i = 1$, \ie, the locations of the landmarks.

{}

\noindent \textbf{Network Input.}
Many works on pose-invariant feature learning take pairs of frontal and profile faces from the same identity as inputs.
Differences between frontal and profile features are penalized to get rid of the pose information.
In contrast, our approach takes one aligned face image, its corresponding landmarks and angles as one unit of input (marked in red in figure~\ref{fig:poseface}).
The angles are estimated from facial landmarks.
Pose-invariant features can be learned in an end-to-end manner by our self-disentanglement approach.

\subsection{Pose-Adaptive ArcFace (PAA) Loss}
Many existing models encounter large performance degradation when recognizing profile faces compared to frontal ones.
Besides intrinsically hard for recognition, another reason is that numbers of frontal and profile samples are highly imbalanced in many training datasets~\cite{cao2018pose}.
In another word, the profile faces are hard and rare samples.

\begin{figure}[htb!]
    \centering
    \includegraphics[width=0.25\textwidth]{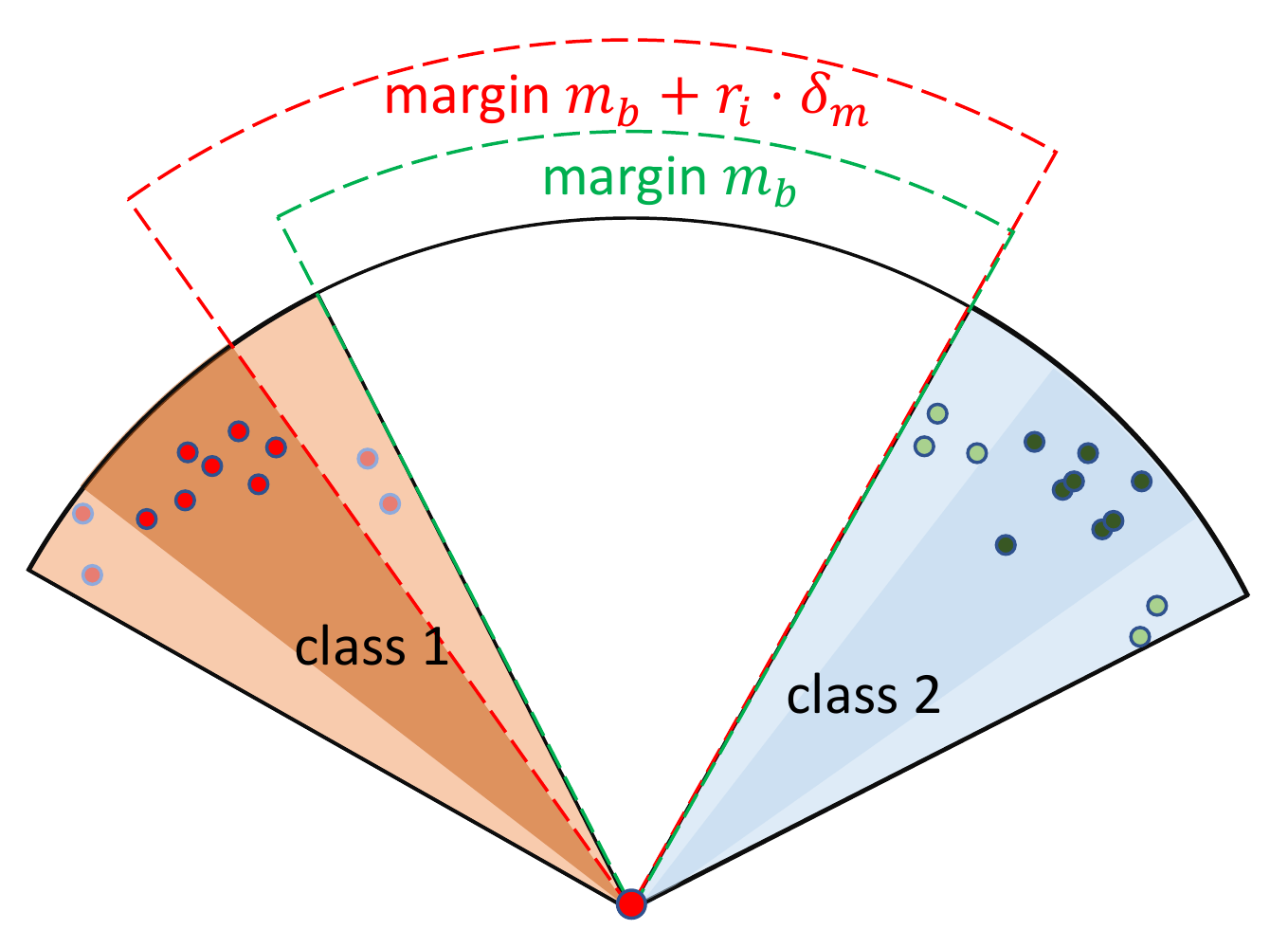}
    \caption{Geometrical interpretation of the adaptive margin. The dots with deeper colors represent profile faces. The harder the samples are, the larger margins will be used.}
    \label{fig:am}
\end{figure}{}

Inspired by the weight-adaptive methods such as  focal loss~\cite{lin2017focal}, Adacos~\cite{zhang2019adacos} and Adaptiveface~\cite{liu2019adaptiveface}, we modify the state-of-the-art ArcFace  loss~\cite{deng2019arcface} by adapting the margin based on the facial angles as shown in figure~\ref{fig:am}.  The margin for sample $i$ is  defined as $m_i = m_b+r_i\cdot \delta_m$.  Here $m_b$ is the base margin and $\delta_m$ is the additional margin.  The amount of added margin  is controlled by ratio $r_i\in [0, 1]$, which is calculated according to the pitch/yaw/roll angles. Profile faces will be assigned with large ratios.  In the end, we formulate the classification loss for identity features as pose-adaptive ArcFace (PAA) loss:

\begin{equation}
  \small
  \begin{split}
    L_{PAA}(r_i)& = -\frac{1}{n}\sum_{i=1}^{N} L_i, \quad \text{where}\\
    & L_i=\log \frac{e^{s\cdot(\cos{(\theta_{y_i}+ m_i))}}}{ e^{s\cdot(\cos{(\theta_{y_i}+m_i))}} + \sum_{j=1, j\neq y_i}^{n}e^{s\cdot \cos{\theta_j}}}.
  \end{split}
    \label{eq:loss}
\end{equation}{}
where 
    $m_i = m_b + r_i \cdot \delta_m$
is the adaptive margin for different poses and other variables such as $\theta_{y_j}, \theta_j, s$ are all consistent with those in  ArcFace~\cite{deng2019arcface}.

\subsection{Pose-Invariant Features}
In our implementation, identity features $F_i$ and pose features $F_p$ are enforced to be distributed in two orthogonal subspaces. By disentangling them, we ensure the generated identity features to be pose-invariant.

\noindent \textbf{Pose-Feature Constraint.}
We treat features $F_p^{L}$ from landmarks as the pseudo labels. The estimated pose features $F_p= \mathbf W_p \cdot F_b$  from the pose module should be consistent with  $F^{L}_p$, \ie,  $\|F^{L}_p - F_p \|_2 = 0$.

\noindent \textbf{Orthogonal Constraint.}
Denote the features from backbone network as $F_b$ and use matrix $\mathbf W_I= [W_I^1, W_I^2, \cdots, W_I^n]$, $\mathbf W_P = [W_P^1, W_P^2, \cdots, W_P^n]$ to represent the transformation of two orthogonal subspaces. Each base of $\mathbf W_I$ and $\mathbf W_P$ are normalized as the scales are not necessary for orthogonal requirement.  We formulate the orthogonal constraint as
\begin{equation}
 \small
\begin{aligned}
    &F_i = \mathbf W_I^T F_b,\quad  F_p = \mathbf W_P^T  F_b, \\
    &\widetilde{\mathbf W}_I = [\frac{W_I^1}{\|W_I^1\|_2}, \frac{W_I^2}{\|W_I^2\|_2}, \cdots, \frac{W_I^n}{\|W_I^n\|_2}], \\
    &\widetilde{\mathbf W}_P = [\frac{W_P^1}{\|W_P^1\|_2}, \frac{W_P^2}{\|W_P^2\|_2}, \cdots, \frac{W_P^n}{\|W_P^n\|_2}], \\
   & \|\widetilde{\mathbf W}_I^T\widetilde{\mathbf W}_P\|_F = 0.
\end{aligned}{}
\end{equation}{}
Here $\|\cdot \|_F$ denotes the Frobenius norm.

\noindent \textbf{PoseFace Loss.} 
Our problem can be formulated as
\begin{equation}
  \small
    \begin{aligned}
      \min\quad  & L_{PAA}(r_i), \\
        s.t. \quad & \|F^{L}_p - F_p \|_2 = 0, \\
         & \|\widetilde{\mathbf W}_I^T\widetilde{\mathbf W}_P\|_F = 0.
    \end{aligned}{}
\end{equation}{}
Introducing Lagrange multipliers $\lambda_{1}, \lambda_{2}$, we define the full version of PoseFace loss as 
\begin{equation}
  \small
    L_{PAA}(r_i) + \lambda_{1} \cdot \|F^{L}_p - F_p \|_2 + \lambda_{2}\cdot  \|\widetilde{\mathbf W}_I^T\widetilde{\mathbf W}_P\|_F.
\end{equation}{}

We call $\lambda_{1} \cdot \|F^{L}_p - F_p \|_2 + \lambda_{2}\cdot  \|\widetilde{\mathbf W}_I^T\widetilde{\mathbf W}_P\|_F$  the Orth loss, which is used to train  pose-invariant features. Our PoseFace loss is a composition of PAA loss and Orth loss.

\section{Experiments}
In this section, we evaluate our PoseFace approach on various face recognition benchmarks with large pose variation. Specifically, we describe the implementation details in section~\ref{sec:exp_details}, ablation study in section~\ref{sec:ablation} and detailed results on benchmarks in section~\ref{sec:exp_bms}. 
In supplementary, we further present sensitive analysis on the hyperparameters $\lambda_1, \lambda_2$ in section A and extra visualizations in sections~C, D.

\subsection{Implementation Details}\label{sec:exp_details}

We use Face Alignment Network (FAN)~\cite{bulat2017far} to generate 68 landmarks and align face images into size $108\times108$ by affine transformations.
Pitch/yaw/roll angles are estimated based on the facial landmarks.
In our implementation, we use a simple function $\frac{|\text{yaw}|}{90^{\circ}}$ to estimate the adaptive ratio $r_i$.
Roll and pitch angles are not considered as the roll's effect will be eliminated by face alignment while face images with large pitch angles are rare~\cite{cao2018pose}.

\begin{figure}[!htb]
  \centering
    \subfloat[]{\includegraphics[width=0.14\textwidth]{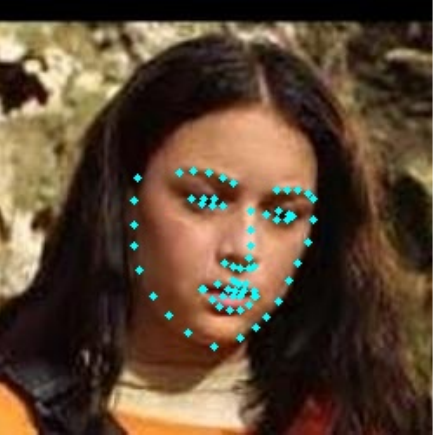}} \quad
    \subfloat[]{\includegraphics[width=0.14\textwidth]{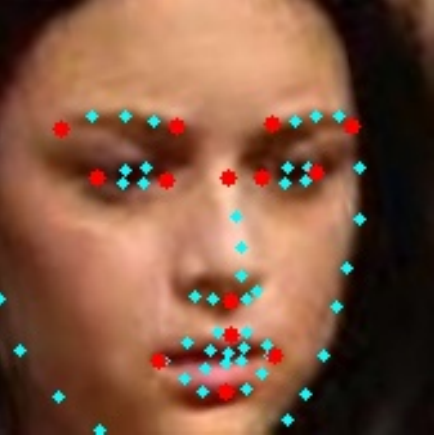}}
    \caption{The procedure of data pre-processing.
      (a) A face image with the detected 68 facial landmarks.
      (b) The corresponding aligned $108\times 108$ face image with the 14 selected landmarks (marked in red) for pose feature extraction.}
    \label{fig:data_example}
\end{figure}

For the AutoEncoder network, we select 14 of 68 landmarks as the input because it is more efficient and some landmarks contain identity information (\eg, the shape of eyes and mouth).
Using all 68 landmarks may degrade the integrity of the identity features.
Figure~\ref{fig:data_example}(b) gives an example of our selected 14 landmarks.
The AutoEncoder Network (see table A1 in supplementary for the detailed structure) is pre-trained by the facial landmarks detected from MS1MV2~\cite{guo2016ms,deng2019arcface} (5.8M images, 85k unique identities).

MS1MV2 is also used as the training set of the PoseFace. Specifically, our PoseFace are trained by setting $\lambda_1=200, \lambda_2=100000, m_b = 0.5$ with various $\delta_m$. Our baselines are trained with $\lambda_1=0, \lambda_2=0, m_b = 0.5$ and $\delta_m=0$, \ie, ArcFace with margin 0.5 with same the network structure. In our experiments, we use ResNet~\cite{he2016deep} with 18, 34 and 50 layers as our backbone networks and rename our models as PoseFace18, PoseFace34 and PoseFace50, respectively. Next we describe our evaluation settings in the involved test databases.

\noindent \textbf{MultiPIE.}
The CMU MultiPIE~\cite{gross2010multi} dataset is the largest multi-view face recognition benchmark and contains 754,204 images of 337 identities from 15 viewpoints and 20 illumination conditions. We follow the evaluation protocol introduced in~\cite{xiong2015conditional}. Images (15 poses, 20 illumination levels) of the first 150 identities are used for training. For testing, one frontal image with neutral illumination is chosen as the gallery image for each of the remaining 100 subjects. The remaining images are used as probes. Due to the limited number of training samples, PoseFace18 is pre-trained on the subset of MS1MV2 (0.68M images, 10K identities) and then fine-tuned on MultiPIE. The rank-1 accuracy is reported with regard to pose for comparison.

\noindent \textbf{CFP.}
Celebrities in Frontal-Profile (CFP) dataset~\cite{sengupta2016frontal} is another challenging dataset with large and unconstrained pose variations. The dataset contains 500 celebrities, each individual has 10 frontal and 4 profile face images. We follow the standard protocol~\cite{sengupta2016frontal} for frontal-profile (FP) face verification. In particular, cross-validation is performed on 10 folders with 350 same and 350 not-same pairs. Means and standard deviations of Accuracy, Equal Error Rate (EER) and Area Under Curve (AUC) over the 10-fold experiments are also reported. PoseFace34 is warmed up by the subset of MS1MV2 for 20 epochs with a learning rate of 0.1. Then it is fine-tuned on the full MS1MV2 for 15 epochs with learning rate $[0.1, 0.01, 0.001]$, 5 epochs for each stage.

\noindent \textbf{CPLFW.}
CPLFW~\cite{zheng2018cross} dataset is a renovation of LFW dataset~\cite{huang2008labeled} with 5,749 identities and 11,652 images. It deliberately searches and selects 3,000 positive face pairs with large pose variations. Negative pairs with the same gender and race are also selected to reduce the influence of attribute difference between positive/negative pairs. PoseFace34 is trained the same way as for CFP. Mean and standard deviation of accuracy are reported under 10-fold cross-validation.

\noindent \textbf{IJB.}
The IJB-B~\cite{whitelam2017iarpa} dataset contains 1,845 subjects with 21.8K still images and 55K frames from 7,011 videos. The IJB-C~\cite{maze2018iarpa} dataset is a further extension of IJB-B and contains about 3,500 identities with a total of 31,334 images and 117,542 unconstrained video frames. PoseFace50 is trained on MS1Mv2 with batch size 512.  The learning rate is initialized as 0.1 and divided by 10 at 5, 10, 15 epochs, and we finish the training at epoch 20. In the 1:1 verification, the numbers of positive/negative matches are 10k/8M in IJB-B and 19k/15M in IJB-C, and we report the TAR on FAR=1e-4, 1e-5 and 1e-6.

\subsection{Ablation Study}\label{sec:ablation}

\noindent \textbf{Effect of our PAA and Orth}. To investigate the effect of our proposed PoseFace, we conduct several ablation experiments to show the superior performances on large-pose face recognition. The PoseFace models are trained by permutations of Orth loss and PAA loss.
The baseline is trained by ArcFace with the same inference structure.
Table~\ref{table:ablation_multipie} summarizes the rank-1 recognition rates.
PoseFace18 achieves significant improvements across all poses, especially for profile poses (more than 60 degrees yaw variation defined as  'profile'~\cite{sengupta2016frontal}).
The rank-1 recognition rates of PoseFace18 with PAA loss outperforms the baseline by 3.69\%, 3.67\% and 2.65\% for $\pm 90^\circ, \pm 75^\circ, \pm 60^\circ$ faces.
If with Orth loss only, the increments are 6.19\%, 6.83\%, 5.00\%.
PoseFace18 achieves the best results with both losses used and the rank-1 recognition rates increase 8.89\%, 8.04\%, 5.80\%.
The experimental results demonstrate the effectiveness of our proposed Orth loss and PAA loss.

\setlength{\tabcolsep}{1pt}
\begin{table}
\begin{center}
  {\footnotesize
\begin{tabular}{lcccccc}
\hline\noalign{\smallskip}
Method  & $\  \pm 90^{\circ}$ \  & $\ \pm 75^{\circ}$ \ & \ $\pm 60^{\circ}$ \ & $\ \pm 45^{\circ}$  \  & $\ \pm 30^{\circ}$  \ & $\ \pm 15^{\circ}$ \  \\
\noalign{\smallskip}
\hline
baseline & 81.68 & 88.03 & 92.75 & 96.99 & 98.77 & 99.37\\
PoseFace18(PAA) & 85.37 &	91.70 &	95.40 &	98.52&	99.36&	99.77 \\
PoseFace18(Orth) & 87.87 &	94.86 &	97.75 &	\textbf{99.85} &	\textbf{99.99} &	99.99 \\
PoseFace18(Orth+PAA) & \textbf{90.58} &	\textbf{96.07} &	\textbf{98.55} &	99.62&	99.97&	\textbf{100} \\
\hline
\end{tabular}
}
\caption{Rank-1 recognition rates (\%) on MultiPIE.}\label{table:ablation_multipie}
\end{center}
\end{table}
\setlength{\tabcolsep}{1.4pt}

\noindent \textbf{Effect of the Landmark Module}. Our Orth Loss is a composition of an orthogonal constraint and a loss for pose features. The orthogonal constraint disentangles the original feature $F_b$ into the identity module and the pose module.
To ensure no other information such as identity flow into the pose module, we design a landmark module to extract features $F_p^L$ which is only related to facial poses, and use $F_p^L$ to supervise the learning of the pose module.
An alternative way is to regress landmark points directly~\cite{peng2017reconstruction} instead of using the landmark module.
We compare these two approaches with/without the orthogonal constraint used and present their performances on IJB benchmarks in table.~\ref{table:abl_ijb}. For fair comparisons, we pick 5 different hyperparameters (from 0.001 to 10, multiply 10 each time) when regressing the landmark points and report the best results.

Without the orthogonal constraint involved, the performances of using landmark module and using landmark points are similar (\eg, TARs are the same when FAR=1e-4 on IJB-C, and with a tiny difference on IJB-B).
However, when introducing the orthogonal constraint, the performance of using landmark points degrades a lot.
Specifically, TAR reduces 0.94\% on IJB-B and 1.03\% on IJB-C when FAR=1e-4.
The possible reason is that when regressing landmarks directly, the learned pose feature $F_p$ cannot guarantee to be pose-related only.
To be more specific, regressing landmark points means $F_p$ can express landmarks, while cannot prevent other information such as identity to be embedded in $F_p$.
This can hurt the recognition performance as valuable identity information is disentangled into the pose module.
Compared to the direct regression, our landmark module builds a one-to-one mapping between pose features and facial landmarks.
Experimentally, the model obtains $0.89\%$ and $0.60\%$ performance boosts at TAR@FAR=1e-4 on IJB-B and IJB-C and that demonstrates the validity of our design.

\setlength{\tabcolsep}{2pt}
\begin{table}[htb!]
  \begin{center}
    {\footnotesize
      \begin{tabular}{l|ccc|ccc}
        \hline
        Method 
        & \multicolumn{3}{c}{IJB-B (TAR@FAR)} & \multicolumn{3}{|c}{IJB-C (TAR@FAR)} \\
         & 1e-5 &  1e-4 & 1e-3 & 1e-5 & 1e-4 & 1e-3 \\
        \hline \hline
        Landmark Points only &  85.74 & 92.65& 95.25 & 90.95& 94.45& 96.51 \\
        Landmark Points +  Orth &  84.66 & 91.71 & 94.68 & 90.91 & 93.42 & 95.85 \\
        \hline
        Landmark Module only &  86.28 & 92.55& 95.26 & 91.31 & 94.45& 96.47 \\
        Landmark Module + Orth &  \textbf{86.61} & \textbf{93.44}&  \textbf{95.92} &   \textbf{92.01} &  \textbf{95.05}&   \textbf{96.97}\\
\hline
      \end{tabular}
    }
  \end{center}
  \caption{Verification evaluation (\%) according to different FARs on IJB-B and IJB-C. Backbone: ResNet50.}      \label{table:abl_ijb}
\end{table}
\setlength{\tabcolsep}{1.4pt}

\begin{figure}[htb!]
    \centering
   \includegraphics[width=0.45\textwidth]{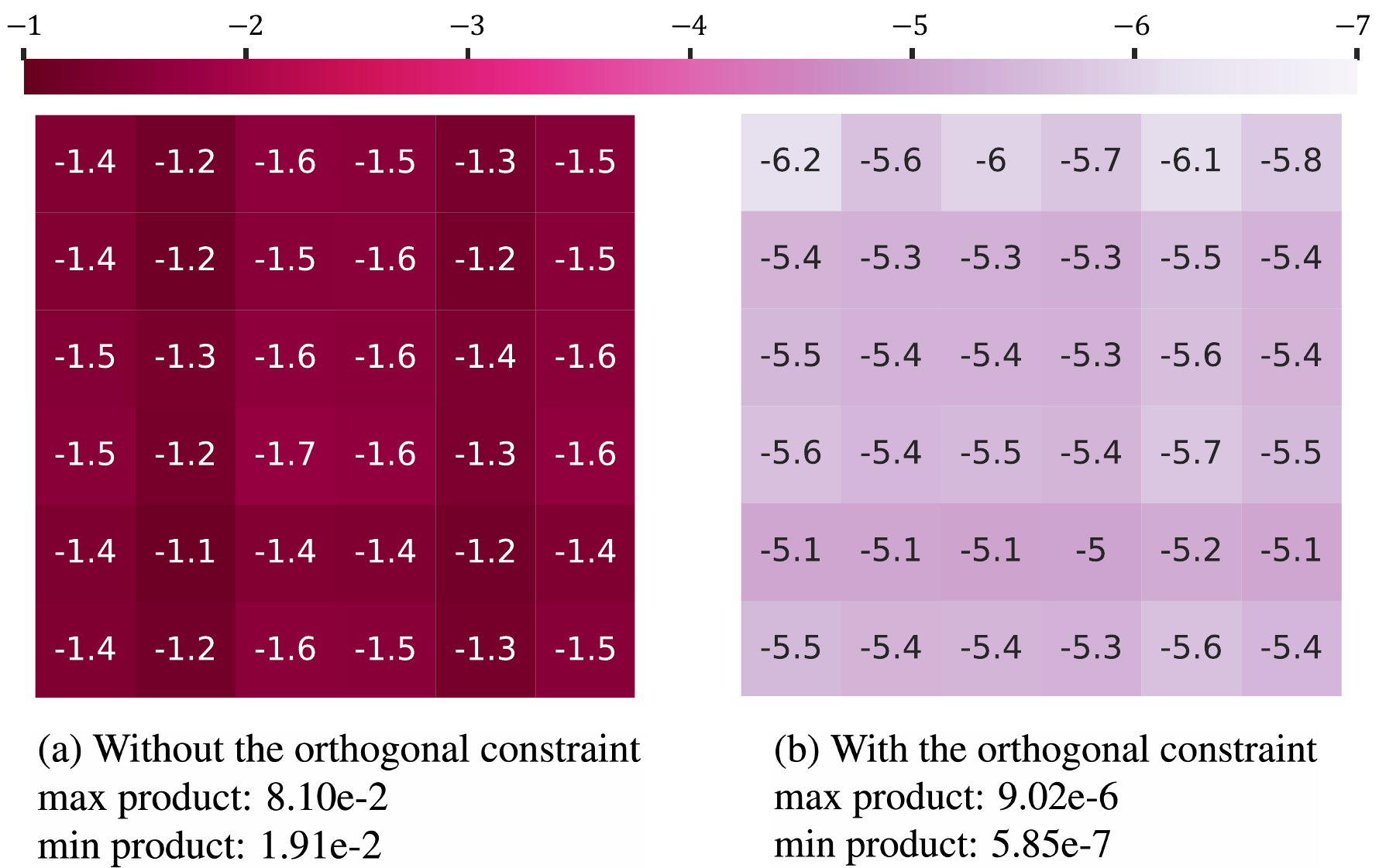}
   \caption{
     We calculate inner products of the mapped identity features and pose features from 10 samples, (a) with and (b) without orthogonal constraints involved.
     Heatmaps show the \textbf{log10} of inner products.
     (See figure~A1 in supplementary for an extended visualization with 100 samples.)
          \textbf{Best viewed in color.}
    }
    \label{fig:ablation_vis}
\end{figure}

\noindent \textbf{Visualization.}
Denote the identity and pose feature for sample $j$ as $F_i^j$ and $F_p^j$.
If distributed in two orthogonal subspaces,  the inner product of an identity feature and a pose feature is always 0 (\ie. $\forall (i, j), \langle F^j_i, F^k_p \rangle =0$).
We select the first 10 test images from MultiPIE and extract  mapped identity  as well as pose features.
All the features are normalized and figure~\ref{fig:ablation_vis} shows the log10 of inner products of all pairs.
With introduce of our orthogonal constraint, the inner products of identity and pose features are significantly decreased by over 4 orders of magnitude.
The largest inner product of an identity and a pose feature is only 9.02e-6 in this example, which indicates the orthogonality between two feature spaces and the efficacy of our disentanglement.

We also visualize the distributions of identity features of frontal and profile faces in figure~A2 (in supplementary).
The visualization further reveals that PoseFace learns more discriminative and pose-invariant features than ArcFace.

{}

\noindent \textbf{Is PoseFace scalable for real-world applications?}
One important question of pose-invariant models is whether it can fit real-world applications.
Recent related works usually need extra computations and labels to frontalize faces~\cite{huang2017beyond,cao2018pose, zhao2018towards}, introduce heavy network modules~\cite{yin2017multi,masi2016pose,xiong2015conditional}, or require pair-wise inputs~\cite{he2019deformable,qian2019unsupervised}.
In contrast, our PoseFace works in an end-to-end manner and scalable for real-world applications.
The introduced self-disentanglement method avoids pair-wise inputs and therefore can be applied to most existing training datasets.
Besides, as introduced modules are all light-weight, PoseFace can be trained with similar configures as ArcFace as illustrated in table~\ref{table:speed}.
Additional GPU memory consumptions are acceptable and the training speed doesn't decrease too much, especially when with large batch sizes.
During inference, PoseFace has no difference with ArcFace as only the identity module is needed.

\setlength{\tabcolsep}{3pt}
\begin{table}[htb!]
  \begin{center}
    \footnotesize
      \begin{tabular}{c|ccccc}
        \hline
        &\multirow{2}{*}{Model} & \multicolumn{4}{c}{Batch Size} \\
        \cline{3-6}
        & & 128 & 256 & 512 & 768 \\
        \hline
        ThroughPut &  ArcFace & 802 & 949 & 986 & 1001 \\
        (images/sec)& PoseFace & 729 & 921 & 956 & 963\\
        \hline
        GPU Memory &  ArcFace & 2599 & 4211 & 6123 & 9309\\
        (MB) & PoseFace & 2645 & 4329 & 6695 & 9798 \\
        \hline                     
      \end{tabular}
      \caption{Training comparison of ArcFace and PoseFace on MS1Mv2 dataset with different batch sizes.
        The experiments are conducted with backbone ResNet50 on 8 1080Ti GPUs.        
              }\label{table:speed}
  \end{center}
\end{table}

\subsection{Results on Benchmarks}\label{sec:exp_bms}
In this section, we present the recognition performances of PoseFace on various benchmarks including MultiPIE, CFP-FP, CPLFW and IJB.
The ``baseline'' in all the following tables indicates a model trained by ArcFace with the same network structure with the corresponding PoseFace.

\noindent
\textbf{Evaluations on MultiPIE.}
We compare our method with the state-of-the-art alternatives on the MultiPIE benchmark in table~\ref{table:multipie}. The c-CNN~\cite{xiong2015conditional} uses dynamically activated kernels to form pose-specific models. DFN-L~\cite{he2019deformable} aligns features across various poses by adding deformable convolutional modules. Other methods listed in the table are all based on GAN. From the results, we can conclude that our approach beats current state-of-the-art works in rank-1 recognition rates. For $\pm 90^\circ$, our recognition rate is 6.51\% higher than the best results reported before (copied from DFN-L~\cite{he2019deformable}). For $\pm 75^\circ$ and $\pm 60^\circ$, the current best results are copied from PIM~\cite{zhao2018towards} with numbers 91.20\% and 97.70\%. Our results are 96.07\% and 98.55\%, respectively. The great performance improvements show the superiority of our proposed method.

\setlength{\tabcolsep}{1pt}
\begin{table}[htb!]
\begin{center}
  {\footnotesize
\begin{tabular}{lcccccc}
\hline\noalign{\smallskip}
Method  & $\  \pm 90^{\circ}$ \  & $\ \pm 75^{\circ}$ \ & \ $\pm 60^{\circ}$ \ & $\ \pm 45^{\circ}$  \  & $\ \pm 30^{\circ}$  \ & $\ \pm 15^{\circ}$ \  \\
\noalign{\smallskip}
\hline
\noalign{\smallskip}
c-CNN~\cite{xiong2015conditional} & 47.26 & 60.66 & 74.38 & 89.02 & 94.05 & 96.97 \\
FNM~\cite{qian2019unsupervised} & 55.80 & 81.30 & 93.70 & 98.20 & 99.50 & \textbf{99.90} \\
TP-GAN~\cite{huang2017beyond} & 64.03 & 84.10 & 92.93 & \textbf{98.58} & \textbf{99.85} & 99.78 \\
DR-GAN~\cite{tran2017disentangled} &  - & - & 86.20 & 90.10 & 94.00 &  97.00 \\
PIM~\cite{zhao2018towards} & 75.00 & \textbf{91.20} & \textbf{97.70} & 98.30 & 99.40 & 99.80 \\
p-CNN~\cite{yin2017multi} & 76.96 & 87.83 & 92.07 & 90.34 & 98.01 &
99.19\\
CAPG-GAN~\cite{hu2018pose} & 77.10 & 87.40 & 93.74 & 98.28 & 99.37 & 99.95 \\
FF-GAN~\cite{yin2017towards} & 61.20 & 77.20 & 85.20 & 89.70 & 92.5 & 94.60 \\
DFN-L\cite{he2019deformable} & \textbf{84.07} & 88.97 & 95.16 & 98.05 &  99.23 & 99.58 \\
\hline
baseline & 81.68 & 88.03 & 92.75 & 96.99 & 98.77 & 99.37\\
PoseFace18(Ours) & \textbf{90.58} &	\textbf{96.07} &	\textbf{98.55} &	\textbf{99.62}&	\textbf{99.97} &		\textbf{100} \\
\hline
\end{tabular}
}
\caption{Comparative performances on MultiPIE benchmark.
  Rank-1 recognition rates (\%)  for different poses are reported. Symbol '-' indicates that the metric is not available.
  Results of other methods are directly copied from the original papers.}\label{table:multipie}
\end{center}
\end{table}
\setlength{\tabcolsep}{1.4pt}

\noindent
\textbf{Evaluations on CFP-FP.}
In table~\ref{table:cfp}, we present the performance of the current state-of-the-art on the CFP benchmark. From the values, we can see that our PoseFace achieves competitive results on the accuracy, EER and AUC. Among the listed methods, DREAM~\cite{cao2018pose} also attempts to extract pose-invariant features by mapping profile features to frontal features. Compared to it, our PoseFace34 achieves better EER (5.09 compared to 6.02, smaller is better) with much fewer layers.
Human performance on verifying the profile faces is listed in the last row of table~\ref{table:cfp}. It can be seen that our PoseFace34 almost achieves a human-level ability to recognize profile faces.

\setlength{\tabcolsep}{2pt}
\begin{table}[htb!]
\begin{center}
{\footnotesize  
\begin{tabular}{lccc}
\hline
Method & Accuracy  & EER &  AUC   \\
\hline
Deep features~\cite{sengupta2016frontal} & 84.91 (1.82) & 14.97 (1.98) & 93.00 (1.55) \\
TPE~\cite{sankaranarayanan2016triplet} & 89.17 (2.35) & 8.85 (0.99) & 97.00 (0.53)\\
DR-GAN~\cite{tran2017disentangled} & 93.41 (1.17) & - (-) & - (-) \\
PIM~\cite{zhao2018towards} & 93.10 (1.01) & 7.69 (1.29) & 97.65 (0.62)\\
p-CNN~\cite{yin2017multi} & \textbf{94.39} (1.17) & \textbf{5.94} (0.11) & \textbf{98.36} (0.05)\\
DREAM18~\cite{cao2018pose} &  - (-) & 7.03 (-) & - (-) \\
DREAM50~\cite{cao2018pose} & - (-) & 6.02 (-) & - (-)\\
\hline
baseline &  	93.09 (1.55)	& 6.83 (1.80) &	97.78 (0.94)\\
PoseFace34(Ours) & \textbf{94.84} (1.18)	& \textbf{5.09} (1.26) &	\textbf{98.82} (0.40) \\
\hline
\hline
\textbf{Human}~\cite{sengupta2016frontal} & 94.57 (1.10) & 5.02 (1.07) & 98.92 (0.46) \\
\hline
\end{tabular}
}
\caption{Comparative performances on the CFP benchmark with the Frontal-Profile protocol designed in the original paper. Results (\%) are the in the format \textit{mean (standard deviation)} over 10-folds in the frontal-profile protocol. Results of other works are directly copied from  original papers.}\label{table:cfp}
\end{center}
\end{table}
\setlength{\tabcolsep}{1.4pt}

\noindent
\textbf{Evaluations on CPLFW.}
We further evaluate our method on the CPLFW benchmark.
Specifically, we re-implement the ArcFace~\cite{deng2019arcface} as the baseline, and Softmax, SV-AM-Sofmax~\cite{wang2018support}, SphereFace~\cite{liu2017sphereface} as the competitors.
The results are shown in table~\ref{table:cplfw}, from which we can see the similar trends that emerged on previous test datasets.
Our PoseFace achieves the best results on the CPLFW benchmark and 
the mean accuracy increases more than 0.55\% compared to the baseline ArcFace.

\setlength{\tabcolsep}{4pt}
\begin{table}[htb!]
\begin{center}
{\footnotesize
\begin{tabular}{ll}
\hline
Method & Accuracy   \\
\hline
Softmax & 85.51 (1.81)\\
SphereFace & 84.66 (2.01)\\
SV-AM-Softmax & 81.84 (2.70)\\
ArcFace & 88.10 (1.95)  \\
PoseFace34(Ours) & \textbf{88.65} (1.97) \\
\hline
\end{tabular}
}
\caption{Mean and standard deviation of the accuracy  (\%) on the CPLFW benchmark under the 10-fold protocol.}\label{table:cplfw}
\end{center}
\end{table}
\setlength{\tabcolsep}{1.4pt}

\noindent
\textbf{Evaluations on IJB-B/IJB-C.}
Besides benchmarks specifically designed for large-pose face recognition, we further evaluate our PoseFace on IJB-B and IJB-C benchmarks, whose images are of large variations on poses, resolutions, occlusions and lighting.
Table~\ref{table:fr_ijb} presents our results.
PoseFace surpasses the implemented baseline on all TAR criteria by $1.48\%$ on FAR=1e-5, $0.81\%$ on FAR=1e-4 and $0.56\%$ on FAR=1e-3 on IJB-B, and by $1.04\%$ on FAR=1e-5, $0.7\%$ on FAR=1e-4, and $0.52\%$ on FAR=1e-3 on IJB-C.

Besides our implemented baseline, we also compare our PoseFace with recent related works.
We pick VGGFace2~\cite{cao2018vggface2}, DCN~\cite{xie2018comparator}, Multicolumn~\cite{xiemulticolumn2018}, PFE~\cite{shi2019probabilistic}, DUL~\cite{chang2020data} and CircleLoss~\cite{sun2020circle} as other baselines,  and present their results in this table.
Note that PFE~\cite{shi2019probabilistic} and DUL~\cite{chang2020data} are two methods specifically designed for low-quality face recognition by introducing the uncertainty learning.
Compared to them, PoseFace50 still achieves better results on IJB-C with fewer layers used (50 compared to 64).
Specifically, PoseFace50 surpasses best results of the two methods by $2.37\%$ on TAR@FAR=1e-5, $0.44\%$ on TAR@FAR=1e-4 and $0.27\%$ on TAR@FAR=1e-3 respectively.
That further shows the superiority of the proposed method.

\setlength{\tabcolsep}{3pt}
\begin{table}[htb!]
  \begin{center}
    {\footnotesize
      \begin{tabular}{l|c|ccc|ccc}
        \hline
        Method & Backbone
        & \multicolumn{3}{c}{IJB-B (TAR@FAR)} & \multicolumn{3}{|c}{IJB-C (TAR@FAR)} \\
        & & 1e-5 &  1e-4 & 1e-3 & 1e-5 & 1e-4 & 1e-3 \\
        \hline \hline
        VGGFace2~\cite{cao2018vggface2} & Res50 & 67.10 & 80.40& 89.10&  74.90 &84.60& 91.30\\
        DCN~\cite{xie2018comparator} &    Res50 &\textbf{84.10} & \textbf{93.00}& - & 88.00 & 94.40 & -\\
        Multicolumn~\cite{xiemulticolumn2018} & Res50 & 70.80 & 83.10& \textbf{90.09} & 77.10 & 86.20 & 92.70\\
        PFE~\cite{shi2019probabilistic} & Res64 & - & - & - & \textbf{89.64} & 93.25 & 95.49 \\
        DUL~\cite{chang2020data} & Res64 & - & - & - & 88.18 & \textbf{94.61} & \textbf{96.70} \\
        CircleLoss~\cite{sun2020circle} & Res100 &-&-&-&89.60&93.95&96.29\\
        \hline
        baseline & Res50 & 85.13 & 92.63& 95.36 & 90.97& 94.35& 96.45 \\
        PoseFace50 & Res50 &  \textbf{86.61} & \textbf{93.44}&  \textbf{95.92} &   \textbf{92.01} &  \textbf{95.05}&   \textbf{96.97}\\
\hline
      \end{tabular}
    }
  \end{center}
  \caption{Verification evaluation (\%) according to different FARs on IJB-B and IJB-C. ``-'' indicates that the author did not report the performance on the corresponding protocols. }      \label{table:fr_ijb}
\end{table}
\setlength{\tabcolsep}{1.4pt}

\section{Conclusions}

In this paper, we propose a novel PoseFace framework to handle the large pose variations.
Specifically, to remove pose information from the identity features, we utilize the facial landmarks as well as an orthogonal constraint to generate the pose-invariant features.
We further propose a pose-adaptive loss to focus training on hard and rare samples (\eg, the profile faces) to address the data imbalance issues.
Extensive experiments on benchmarks present compelling results, which show the superiority of our PoseFace.
As a general framework, PoseFace can be further extended with other classification losses besides ArcFace as well as metric learning losses (\eg, triplet loss~\cite{schroff2015facenet} and N-pair loss~\cite{sohn2016improved}).
Besides, the idea of combining AutoEncoder and orthogonal spaces can also be applied to other fields to disentangle irrelevant features for specific tasks.

{\small
\bibliographystyle{ieee_fullname}
\bibliography{egbib}
}
\appendix
\section{Ablation Study on Parameters $\lambda_1, \lambda_2$}
In our work, $\lambda_1, \lambda_2$ in Eq.(5) are two hyperparameters for the Orth loss, where $\lambda_1$ amplifies the errors of predicted pose features while $\lambda_2$ controls the orthogonality.
We first let the $\lambda_1=100$ and report recognition results on MultiPIE with different $\lambda_2$ as shown in table~\ref{table:ablation_multipie_2}.
When no orthogonal constraints added, the model performs poorly especially for large pose faces.
For example, the recognition rate is only $80.87\%$ and $88.98\%$ for faces with yaws $\pm 90^{\circ},  \pm 75^{\circ}$.
The performances increase significantly when enlarging $\lambda_2$.
When $\lambda_2=1e5$, the model achieves overall the best results on the benchmark.
Further increase the value exponentially does not hurt the performances too much.
Specifically, setting $\lambda_2=1e6$ still achieves a $6.1\%, 5.24\%$ boosts for $ \pm 90^{\circ},  \pm 75^{\circ}$, compared to no orthogonal penalty added.

\setlength{\tabcolsep}{1.4pt}
\begin{table}[htb!]
\begin{center}
  {\small
\begin{tabular}{ccccccc}
\hline
$\lambda_2$  & $\  \pm 90^{\circ}$ \  & $\ \pm 75^{\circ}$ \ & \ $\pm 60^{\circ}$ \ & $\ \pm 45^{\circ}$  \  & $\ \pm 30^{\circ}$  \ & $\ \pm 15^{\circ}$ \  \\
  \hline
  0 & 80.87 & 88.98 & 93.51 & 97.08 & 98.43 &  99.48 \\
  1e3 & 82.04 & 90.77 & 94.72 & 97.56 & 99.27 & 99.52 \\
  1e4 & 83.49 & 92.22 & 95.85 & 98.79 & 88.69 & 99.87 \\
  1e5 & \textbf{87.58} & \textbf{95.41} & \textbf{97.99} & 99.43 & \textbf{99.98} & \textbf{100.00} \\
  1e6 & 86.97 & 94.22 & 97.87 & \textbf{99.58} & 99.92 & 99.99 \\
\hline
\end{tabular}
}
\caption{Rank-1 recognition rates (\%) on MultiPIE for different poses. $\lambda_1$ is fixed to be 100.}\label{table:ablation_multipie_2}
\end{center}
\end{table}

We further let $\lambda_2=1e5$ and train models with different $\lambda_1$.
Similar phenomenon observed as indicated in table~\ref{table:ablation_multipie_1}.
A small $\lambda_1$ (\eg, 10 in the table) means the disentangled pose features may be incorrect and contain identity features.
The performances are consistently high if when $\lambda_1\geq 100$ in the table,
while over-emphasizing the loss term can also hurt recognition.
The best $\lambda_1$ are around 100 or 200.

According to the results on MultiPIE, we then empirically set $\lambda_1=200, \lambda_2=1e5$ in all experiments presented in the paper.

\setlength{\tabcolsep}{1.4pt}
\begin{table}[htb!]
\begin{center}
  {\small
\begin{tabular}{l|cccccc}
\hline
$\lambda_1$  & $\  \pm 90^{\circ}$ \  & $\ \pm 75^{\circ}$ \ & \ $\pm 60^{\circ}$ \ & $\ \pm 45^{\circ}$  \  & $\ \pm 30^{\circ}$  \ & $\ \pm 15^{\circ}$ \  \\
\hline
  10 & 81.24 & 88.94 & 92.64 & 97.31 & 99.00 & 99.62 \\
  100 & 87.58 & \textbf{95.41} & \textbf{97.99} & 99.43 & 99.98 & \textbf{100.00} \\
  200 & \textbf{87.87} & 94.86 & 97.75 & \textbf{99.85} &\textbf{99.99} & 99.99 \\  
  500 & 86.93 & 95.06 & 97.89 & 99.61 & 99.88 & 99.99 \\
  1000 & 86.82 & 93.48 & 97.53 & 99.39 & 99.87 & 99.99 \\
\hline
\end{tabular}
}
\caption{Rank-1 recognition rates (\%) on MultiPIE for different poses. $\lambda_2$ is fixed to be 1e5.}\label{table:ablation_multipie_1}
\end{center}
\end{table}

\section{Auto-Encoder Structure}
 The structure of our Auto-Encoder network is summarized in Table~\ref{table:ae}. The encoder is a sequence of 5 convolution blocks, consisting of a convolution layer, batch normalization layer, ReLU and max pooling. A fully connected layer is applied to the end of the encoder to output the final feature vector. The decoder is exactly the inverse version of the encoder. 

\begin{table}[htb!]
\begin{center}
  {\footnotesize
\begin{tabular}{l}
  \hline\hline
  Encoder \\
  \hline
  $[conv, 3\times 3, 32] , BN, ReLU, pooling $\\
  $[conv, 3\times 3, 64] , BN, ReLU, pooling $ \\
  $[conv, 3\times 3, 128] , BN, ReLU, pooling $  \\
  $[conv, 3\times 3, 256] , BN, ReLU, pooling $\\
  $[conv, 3\times 3, 512] , BN, ReLU, pooling $ \\
  $[fc, 512], l2\ norm$  \\
  \hline\hline
  Decoder \\
  \hline
  $[fc, 512\times 4\times 4] , reshape$ \\
  $[upsample, 7\times7], [conv, 3\times3, 256], BN, ReLU$ \\
  $[upsample, 14\times14], [conv, 3\times3, 128], BN, ReLU$ \\
  $[upsample, 28\times28], [conv, 3\times3, 64], BN, ReLU$\\
  $[upsample, 56\times56], [conv, 3\times3, 32], BN, ReLU$ \\
  $[upsample, 108\times 108], [conv, 3 \times 3, 14],  sigmoid $\\
  \hline
\end{tabular}
}
\caption{Details of our Auto-Encoder structure}\label{table:ae}
\end{center}
\end{table}

\begin{figure*}[t!]
    \centering
    \subfloat[Without Orthogonality \\ max product: 9.12e-2 \\ min product: 2.41e-5]{
      \includegraphics[trim=0 50 0 0,clip,width=0.5\textwidth]{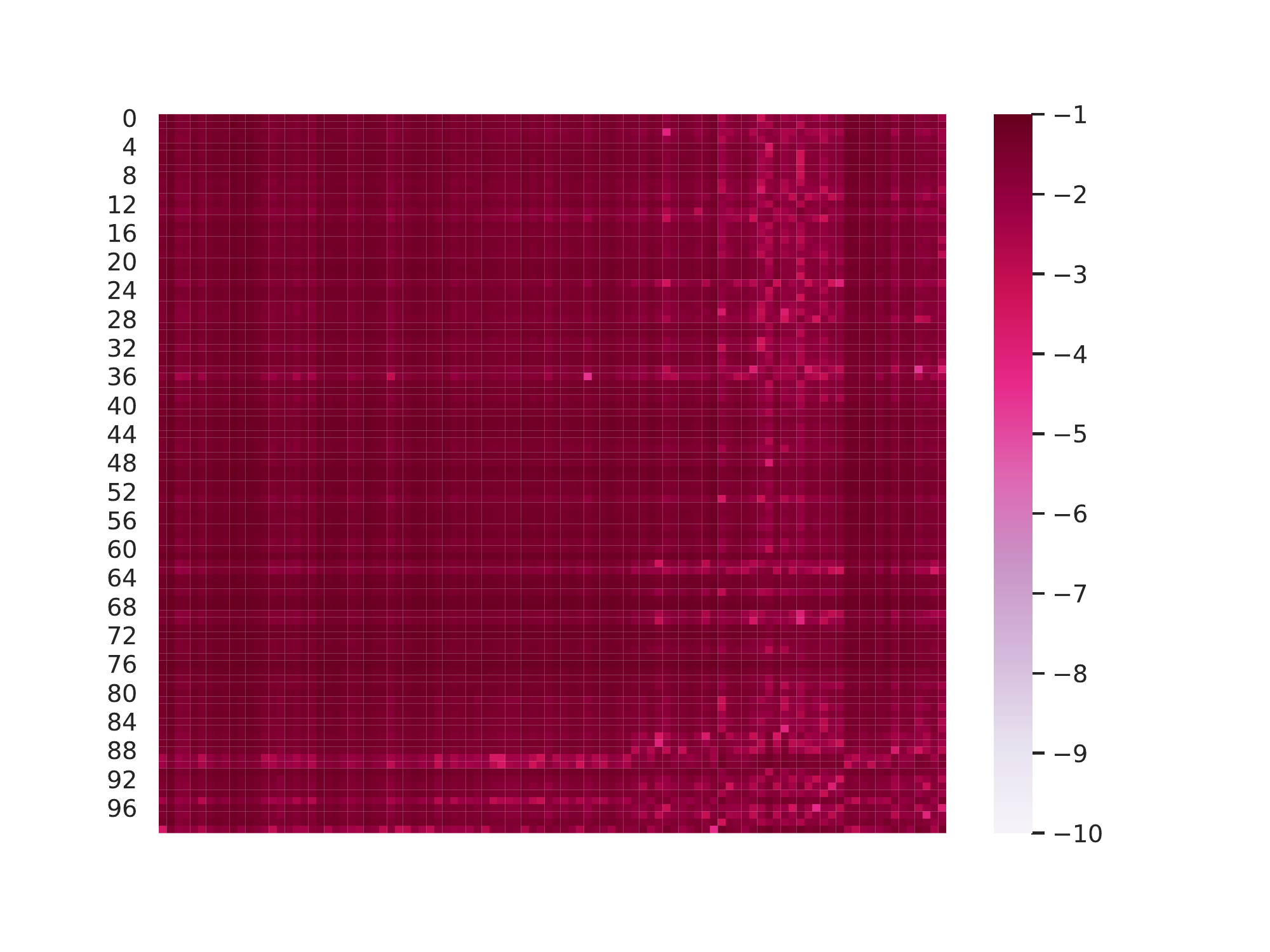}
    }
    \subfloat[With Orthogonality \\max product: 2.14e-5 \\ min product: 3.11e-10]{
      \includegraphics[trim=0 50 0 0,clip, width=0.5\textwidth]{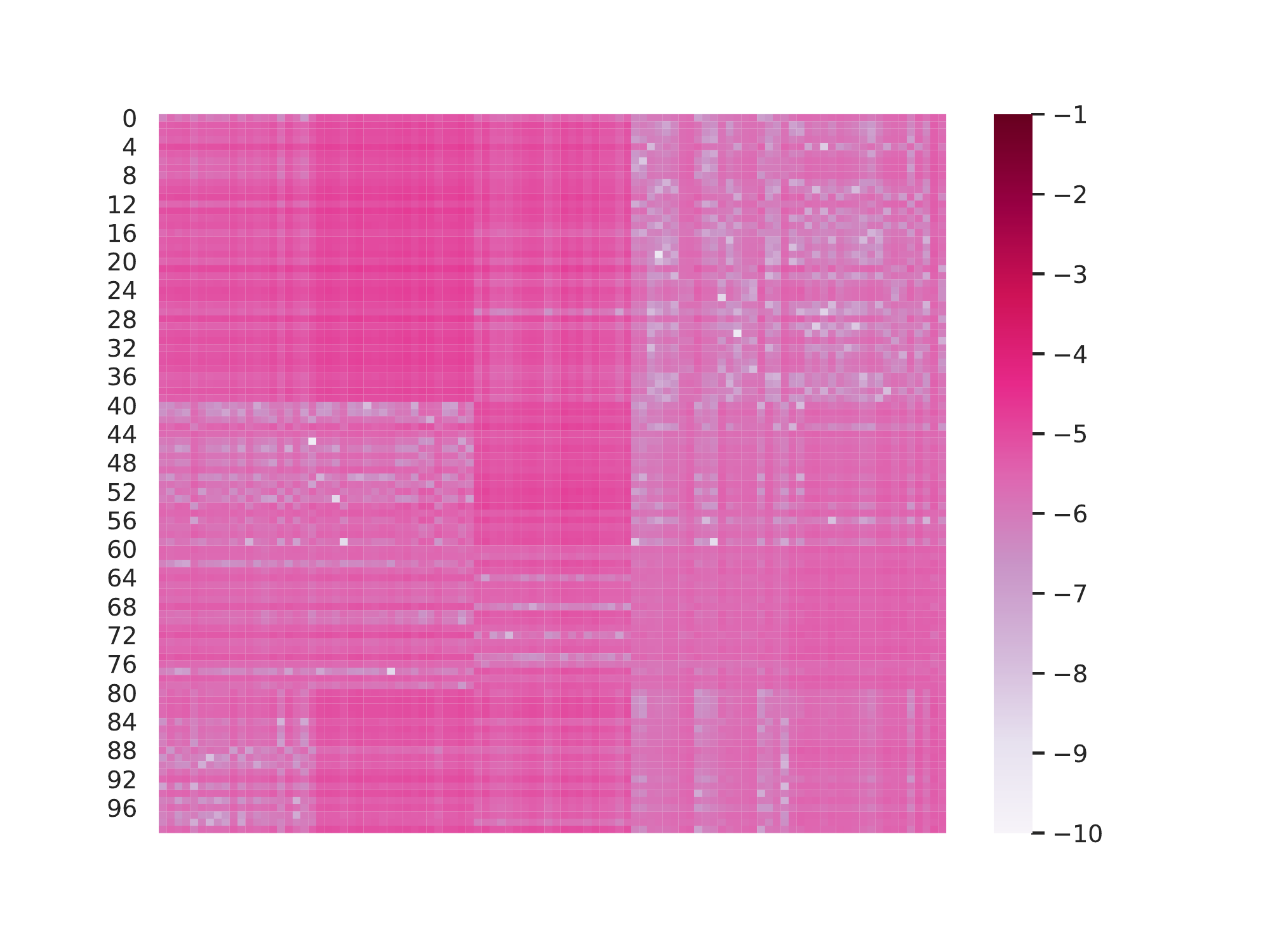}
    }
    \caption{Extended visualization of figure~6.
       \textbf{Best viewed in color.}
    }
    \label{fig:ablation_vis_ext}
  \end{figure*}{}  
  
\section{Extended Visualization of Figure~6}
Figure~\ref{fig:ablation_vis_ext} presents a extended visualization of figure~6 with 100 samples used.
The inner products of a pose feature and an identity feature is consistently of a small order of magnitude.
That indicates the orthogonality between two feature spaces and demonstrates the efficacy of our disentanglement.

\section{Visualization of Feature Distributions}

\begin{figure}[htb!]
    \centering
        \subfloat[Features from ArcFace\\avg. intra-class distance: 5.3e-3 \\ avg. inter-class distance: 1.33]{\includegraphics[width=0.22\textwidth]{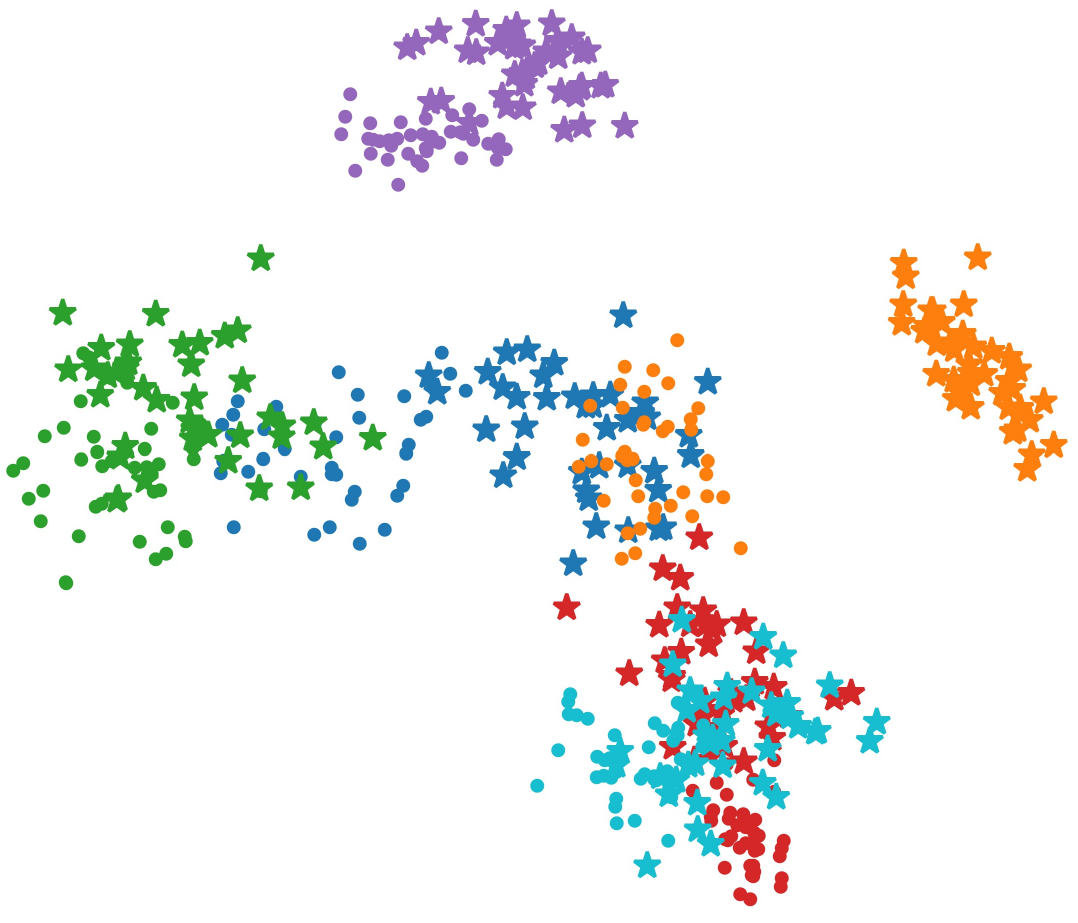}}\quad
    \subfloat[Features from PoseFace \\avg. intra-class distance: 4.3e-3 \\ avg. inter-class distance: 1.39 ]{\includegraphics[width=0.22\textwidth]{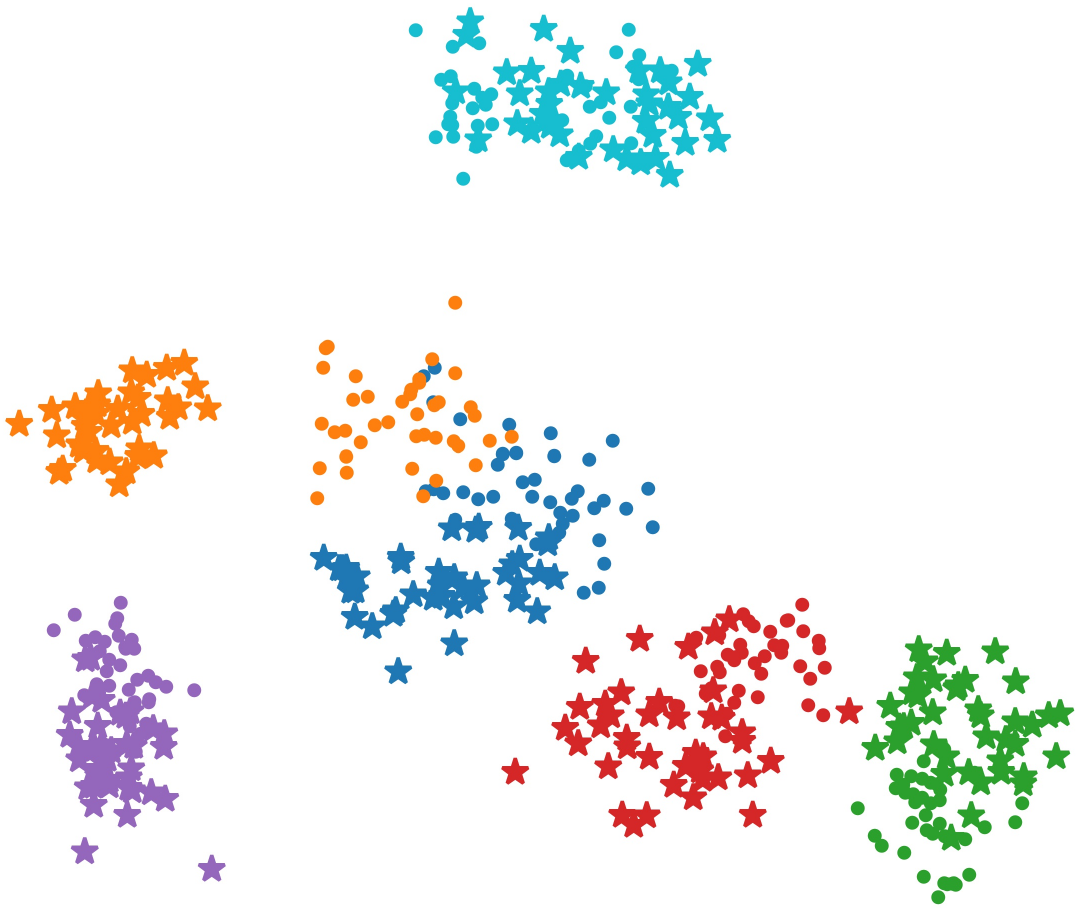}}
    \caption{Visualization of the features from baseline (ArcFace) and PoseFace.
      Each color denotes an identity and each identity contains 40 frontal (yaw angle $0^\circ$, marked in dots) and 40 profile faces (yaw angle $90^\circ$, marked in stars). The features are projected to the top 2 dimensions via PCA.
       \textbf{Best viewed in color.}
    }
    \label{fig:ablation}
  \end{figure}{}

We visualize deep feature distributions of the baseline (ArcFace) and our PoseFace on MultiPIE in figure~\ref{fig:ablation}. Compared to ArcFace, features of our PoseFace are more discriminative in pose variations cases clearly indicated by the figure.
We further calculate the average intra-class and inter-class distances.
Compared to ArcFace, PoseFace decreases the average inter-class distance from 0.0053 to 0.0043 while increases the average intra-class distance from 1.33 to 1.39.
In addition, for each identity of PoseFace, the profile face features (marked in stars) are clustered with the corresponding frontal face features (marked in dots) more tightly than that of in ArcFace.
To sum up, the visualization reveals the proposed PoseFace learns more discriminative and pose-invariant features than ArcFace.

\end{document}